\documentclass[letterpaper]{article} 
\usepackage[submission]{aaai25}  
\usepackage{times}  
\usepackage{helvet}  
\usepackage{courier}  
\usepackage[hyphens]{url}  
\usepackage{graphicx} 
\usepackage{natbib}  
\usepackage{caption} 
\usepackage{algorithm}
\usepackage{algorithmic}

\usepackage{newfloat}
\usepackage{listings}
\DeclareCaptionStyle{ruled}{labelfont=normalfont,labelsep=colon,strut=off} 
\floatstyle{ruled}
\newfloat{listing}{tb}{lst}{}
\floatname{listing}{Listing}
\usepackage{xcolor}

\title{Bi-Directional Mental Model Reconciliation for\\ Human-Robot Interaction with Large Language Models}  

\author{Nina Moorman, Michelle Zhao, Matthew B. Luebbers, Sanne Van Waveren, Reid Simmons, Henny Admoni, Sonia Chernova, and Matthew Gombolay}

\begin{document}

\maketitle

\begin{abstract}
In human-robot interactions, human and robot agents maintain internal mental models of their environment, their shared task, and each other. The accuracy of these representations depends on each agent's ability to perform theory of mind, i.e. to understand the knowledge, preferences, and intentions of their teammate. When mental models diverge to the extent that it affects task execution, reconciliation becomes necessary to prevent the degradation of interaction. We propose a framework for bi-directional mental model reconciliation, leveraging large language models to facilitate alignment through semi-structured natural language dialogue. Our framework relaxes the assumption of prior model reconciliation work that either the human or robot agent begins with a correct model for the other agent to align to. Through our framework, both humans and robots are able to identify and communicate missing task-relevant context during interaction, iteratively progressing toward a shared mental model.

\end{abstract}

\section{Introduction} 

Mental models are abstract representations of reality, used for reasoning about cause and effect, and for making decisions in an individual's environment \cite{wilson1989mental}. Though the term originates from human psychology, it can also be applied to robotic agents to describe their formalized world and task models, programmed to support autonomous decision-making \cite{tabrez2020survey}. Prior work in human factors has shown that the degree of mental model synchronization between collaborators on a task is correlated with team performance \cite{mathieu2000influence}. To achieve this synchronization, humans rely on their theory of mind capacity to infer the mental models of their teammates through observation, communicating when disagreements are identified \cite{andrews2023role}. To achieve fluent human-robot teaming, we must develop systems with a similar capacity for identifying and reconciling mental model discrepancies during interaction.

\begin{figure}
    \centering
    \includegraphics[width=.9\linewidth]{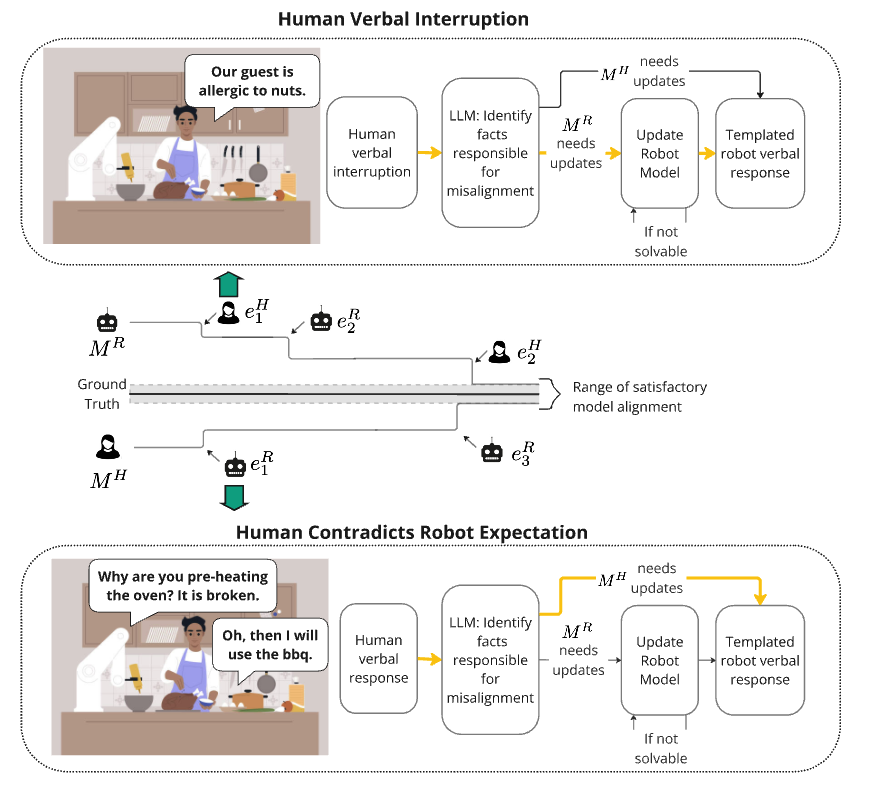}
    \caption{In our pipeline, the robot and human can prompt mental model reconciliation via natural language.} 
    \label{fig:overview}
\end{figure}


Prior human-robot model reconciliation methods have typically been uni-directional: either a robot's model is aligned with an expert human's model (e.g., in learning from demonstration~\cite{argall2009survey}), or a human's model is aligned with an expert robot's model (e.g., in autonomous decision support or behavior elicitation/coaching \cite{tabrez2019explanation, sreedharan2021foundations}). However, in real-world human-robot interactions, the diversity of environments and users means neither the human nor the robot is likely to start with a complete mental model for the task. 

We propose a framework for bi-directional mental model reconciliation between human and robotic agents. The framework facilitates iterative updates of both human and robot models through semi-structured natural language dialogue, initiated either by verbal interruptions from the human or upon the observation of human actions that contradict the robot's expectation. This iterative process allows both humans and robots to share knowledge and preferences during the interaction, and gradually form a shared, mutually satisfactory mental model for the task.

The proposed contribution of our work is the following:
\begin{enumerate}
    \item A theoretical framework for bi-directional human-robot mental model reconciliation.
    \item An instantiation of that framework which represents the robot's model via Planning Domain Definition Language (PDDL), represents shared mental model context as structured facts \cite{knepper2017implicit}, and leverages a large language model (LLM) to process natural language dialogue between the human and robot agents.
    \item A human-subjects experiment evaluating the performance of the proposed method for facilitating iterative model updates via natural language communication.
\end{enumerate}

\section{Methodology}
\subsubsection{Problem Formulation}
In our setup, a human-robot team shares a collective task, specified within a ground-truth task context, $c^{GT}$. In practice, $c^{GT}$ comprises knowledge involving the task, environment, and each agent's capabilities and preferences, such that the task can be completed to each agent's satisfaction. Neither agent is assumed to fully know $c^{GT}$; instead, each begins with their own understanding of the context, $c^R_0$ and $c^H_0$. 

The robot and human mental models, $M^R$ and $M^H$, combine each agent's current context with a decision-making capacity. Throughout the interaction, $M^R$ yields both a policy for the robot to follow $\pi^R$, and a prediction of the human's policy $\pi^{R(H)}$. Likewise, $M^H$ yields a human policy $\pi^H$ and predicted robot policy $\pi^{H(R)}$.

The solution to the bidirectional model reconciliation problem is a set of explanations $E^R \cup E^H = \{e^R_1, ...e^R_n\} \cup \{e^H_1, ...e^H_m\}$, that minimizes $d(\pi^{H(R)}, \pi^R) + d(\pi^{R(H)}, \pi^H)$, with each explanation aimed at communicating missing contextual information to the other agent, thus updating that agent's mental model. The reconciliation is deemed complete when $d(\pi^{H(R)}, \pi^R) < \epsilon$, and $d(\pi^{R(H)}, \pi^H) < \epsilon$. 

\subsubsection{Research Questions}
In this work, we investigate the following research questions.
\begin{enumerate}
    \item \textbf{RQ1)} As a function of the number of iterations, how does bidirectional model reconciliation impact the accuracy of the robot's and the human's mental model, as compared to ground truth?
    \item \textbf{RQ2)} As a function of the number of iterations, how does bidirectional model reconciliation impact the alignment between the robot's and the human's mental model? 
    \item \textbf{RQ3)} As a function of the number of iterations, how does bidirectional model reconciliation impact user attitudes towards and perceptions of the robot?
\end{enumerate} 

\subsubsection{Approach}
Our proposed approach is depicted in Figure \ref{fig:overview}. To evaluate our framework, we implement the robot mental model $M^R$ using a common planning language (PDDL \cite{aeronautiques1998pddl}); solving the planning problem affords $\pi^R$ and $\pi^{R(H)}$. The human mental model $M^H$ represents the human's internal decision-making. To facilitate the alignment of task-relevant context, we represent $c^R_t$ and $c^H_t$ as sets of facts (fact-based models) that reflect knowledge believed by an agent, similar to \citet{knepper2017implicit}.

Given their initial fact-based model contexts, the human and robot formulate their respective plans and begin executing them concurrently. Model reconciliation is initiated in two ways: (1) when the human interrupts with a verbal utterance and (2) when the robot notices a deviation from expected human behavior ($\pi^{R(H)} \neq \pi^H$). In this second case, the robot provides a templated verbal interruption that communicates the anticipated and actual human behavior, asking the human to clarify the discrepancy. 

Upon receiving either the interruption or the clarification from the human, the pipeline employs an LLM to input the human's utterance, and output whether the robot or human contexts are missing information, and what fact(s) could be added to either to rectify the discrepancy. If the robot's context has been updated, another LLM takes the new $c^R_t$, and returns an updated robot mental model $M^R$. Once updated, the robot provides a templated verbal explanation of the update. On the other hand, if the human's context has been updated, the robot provides the human with a templated verbal explanation of the new fact(s). Finally, the human is asked to restate what the robot has indicated, ensuring mutual understanding of the respective model updates.

\subsubsection{Proposed Evaluation}
We propose a human subject experiment to evaluate the accuracy of and alignment between the robot and human mental model, and to investigate the resulting user perceptions of and attitudes toward the robot. After obtaining participants' consent and demographics, the human and robot are each given an initial mental model. In this work, we conduct mental model reconciliation in the cases where both mental models contain correct but incomplete information. To accomplish the collaborative task, the human and robot must identify when their mental models lack information, prompt the other agent, and exchange the missing information.

We define the ground truth mental model as the union of the facts initially given to the robot and the human. To evaluate mental model accuracy we report the edit distance\footnote{We define edit distance here as the number of facts that would need to be edited such that the two mental models are the same.} between the ground truth mental model and the final human mental model. To evaluate the alignment between the robot and human mental models, we report the edit distance between the two fact-based models, and visualize the changes in edit distance over time. 

Our evaluation domain involves organizing and hosting a dinner party, with tasks such as picking a dish, cooking, setting the table, and loading the dishwasher. We propose to evaluate our mental model reconciliation system in scenarios where either, both, or neither models have missing information. 
At the end of each task, a post-task questionnaire is administered that measures the human's perceived task success, and the human's mental model using the Situation Awareness Global Assessment Technique (SAGAT) \cite{endsley1988situation} over the content of the fact-based model. At the end of the study, we administer a questionnaire that measures the human's perceptions of and attitudes toward the robot, including perceived workload \cite{hart1986nasa}, acceptance \cite{belanche2012integrating}, and trust \cite{jian2000foundations}.




\newpage
\bibliography{aaai25}
\end{document}